\documentclass[letterpaper,10pt,journal,twoside]{IEEEtran}
\usepackage{amsmath,amssymb,amsfonts}
\usepackage{bm}

\usepackage{eucal}
\usepackage{pgfplots}
\pgfplotsset{compat=1.18} 
\usepackage{graphicx} 
\usepackage{textcomp}
\usepackage{xcolor}
\usepackage[hidelinks]{hyperref}
\usepackage{verbatim}
\usepackage{algorithm}
\usepackage[noend]{algpseudocode}
\usepackage{multirow}
\usepackage{booktabs}
\usepackage{tabularx}
\usepackage{gensymb}
\usepackage{lipsum}

\usepackage{soul}
\usepackage{mathptmx, mathtools}
\usepackage[utf8]{inputenc}
\usepackage[nodisplayskipstretch]{setspace}
\usepackage{cite}
\usepackage{multicol}
\usepackage{listings}
\usepackage{tcolorbox}
\usepackage{placeins} 

\usepackage{array}
\usepackage{etoolbox}

\makeatletter

\def\ps@IEEEtitlepagestyle{%
    \def\@oddhead{%
        \footnotesize
        IEEE ROBOTICS AND AUTOMATION LETTERS. PREPRINT VERSION. ACCEPTED SEPTEMBER, 2026
        \hfill \thepage}
    \def\@oddfoot{}
}

\makeatother

\begin{document}

\title{Vision-Language Grounded Task-Context-Aware Imitation Learning for Robotic Disassembly}

\author{
Jeon Ho Kang$^{1}$, Igal Tamarkin$^{1}$, Ethan Niu$^{1}$, Ian Novales$^{1}$, and Satyandra K. Gupta$^{1}$  
\thanks{\textcopyright2026 IEEE. Personal use of this material is permitted. Permission from
IEEE must be obtained for all other uses, in any current or future media,
including reprinting/republishing this material for advertising or promotional
purposes, creating new collective works, for resale or redistribution to servers or lists, or reuse of any copyrighted component of this work in other works.}
\thanks{Manuscript received: May 15, 2026; Revised: July 31, 2026; Accepted: September 3, 2026.}
\thanks{This paper was recommended for publication by Editor Pascal Vasseur upon evaluation of the Associate Editor and Reviewers' comments.}  
\thanks{$^{1}$ J.H. Kang, I. Tamarkin, E. Niu, I. Novales, and S.K. Gupta are with the Viterbi School of Engineering, University of Southern California, Los Angeles, USA.  
{\tt\footnotesize \{jeonhoka, guptask\}@usc.edu}}  
\thanks{Digital Object Identifier (DOI): see top of this page.}
}

\markboth{IEEE Robotics and Automation Letters. Preprint Version. Accepted September, 2026}
{Kang \MakeLowercase{\textit{et al.}}: Task-Context-Aware Imitation Learning for Robotic Disassembly}

\maketitle

\thispagestyle{IEEEtitlepagestyle}

\begin{abstract}
Real-world robotic disassembly requires long-horizon execution, where robots must perform ordered sequences of manipulation tasks across multiple parts within a single scene. Multiple valid task goals and diverse assembly configurations make it difficult for imitation policies to infer the intended skill from raw observations alone, particularly when training data cannot cover the combinatorial diversity of real-world configurations and part geometries. We show that incorporating task context through language alleviates these challenges by providing explicit structure for skill selection and associating language-specified tasks with their corresponding manipulation targets in the visual scene. The proposed framework combines hierarchical task selection with task-context-aware imitation learning to ground language instructions in spatial visual representations for robotic disassembly. The resulting framework generalizes across diverse connector geometries and assembly configurations without requiring explicit object annotations. Our method improves end-to-end task success by 35 percentage points over the baseline diffusion policy and by 75 percentage points over the previous task-context-aware baseline. Demonstration videos and code are available at: 
\url{https://rros-lab.github.io/task-context-aware-il-with-vlm.github.io/}
\end{abstract}


\section{Introduction}

\noindent
\IEEEPARstart{T}{he} robotic disassembly of real-world products involves multiple parts, each requiring distinct manipulation skills, and these behaviors are far more complex than pick-and-place operations. Learning from human demonstrations offers an effective path to acquire such skills. In disassembly settings, however, a single scene may contain multiple objects requiring different skills, making goal and skill selection a major challenge for imitation learning. When relying solely on raw visual and force observations, the required data grows combinatorially with the number of possible configurations. A second challenge is that long-horizon demonstrations often contain idle motions around key subtasks. Behavior cloning absorbs these inactive segments into the policy, causing temporal freezing during rollout, where the robot remains idle instead of transitioning to the next subtask. Although pauses can be reduced during data collection or manually filtered, doing so requires costly refinement and may remove semantically meaningful waiting behavior, such as pauses for human intervention or task completion.

Previous work \cite{kangtaskcontext2025} has demonstrated that explicit task context switching using natural language mitigates the \textit{temporal freezing}. However, its evaluation relies on human-provided online subtask labels, and the proposed CNN classifier was not evaluated for autonomous task selection. Our method performs automatic skill labeling and online subtask selection during execution, eliminating the need for human annotation in end-to-end imitation learning. Moreover, the evaluation in \cite{kangtaskcontext2025} is limited to a fixed assembly setup (NIST board), without assessing generalization to unseen configurations and objects. Changing the assembly configuration or introducing unseen parts can cause the policy to rely on spurious correlations between object appearance and scene layout rather than the intended language command, leading to incorrect target selection and poor generalization.

To address these challenges, we introduce a task-context-aware policy that grounds language in spatial visual representations without requiring object-level annotations, using a counterfactual loss to reinforce task-relevant language–object associations and enable robust target disambiguation across held-out assembly configurations. Additionally, we propose a high-level Vision--Language Model (VLM)-based planner for semantic task selection and generalization to held-out connector mechanisms using the existing skill library.

The proposed framework is evaluated on contact-rich disassembly tasks in which mechanically coupled components are separated by overcoming retention forces arising from friction, interference fits, magnetic attraction, or similar mechanisms. These tasks require sequencing manipulation primitives to progressively release mechanical constraints. Accordingly, connector disassembly serves as a representative application domain, while held-out mechanisms and assemblies demonstrate generalization to previously unseen disassembly tasks that can be addressed using the same manipulation skill library. The framework is designed to accommodate additional manipulation primitives as the skill library expands.
Our contributions include the following:

\begin{enumerate}

\item We introduce a task-context-aware imitation learning framework that combines task language, visual representations, and counterfactual supervision to ground task-relevant target objects without explicit object annotations, including in held-out assembly configurations.

\item We propose a hierarchical VLM-based task-selection framework that enables semantic task selection for held-out assembly configurations and connector mechanisms using the existing skill library.

\item We evaluate the proposed framework across a diverse set of connector assembly configurations with varying types, spatial arrangements, and held-out parts, extending beyond the evaluation settings considered in \cite{kangtaskcontext2025}.
\vspace{-0.7em}
\end{enumerate}

\begin{figure*}[tp]
    \centering
    \smallskip
    \includegraphics[width=0.8\linewidth]{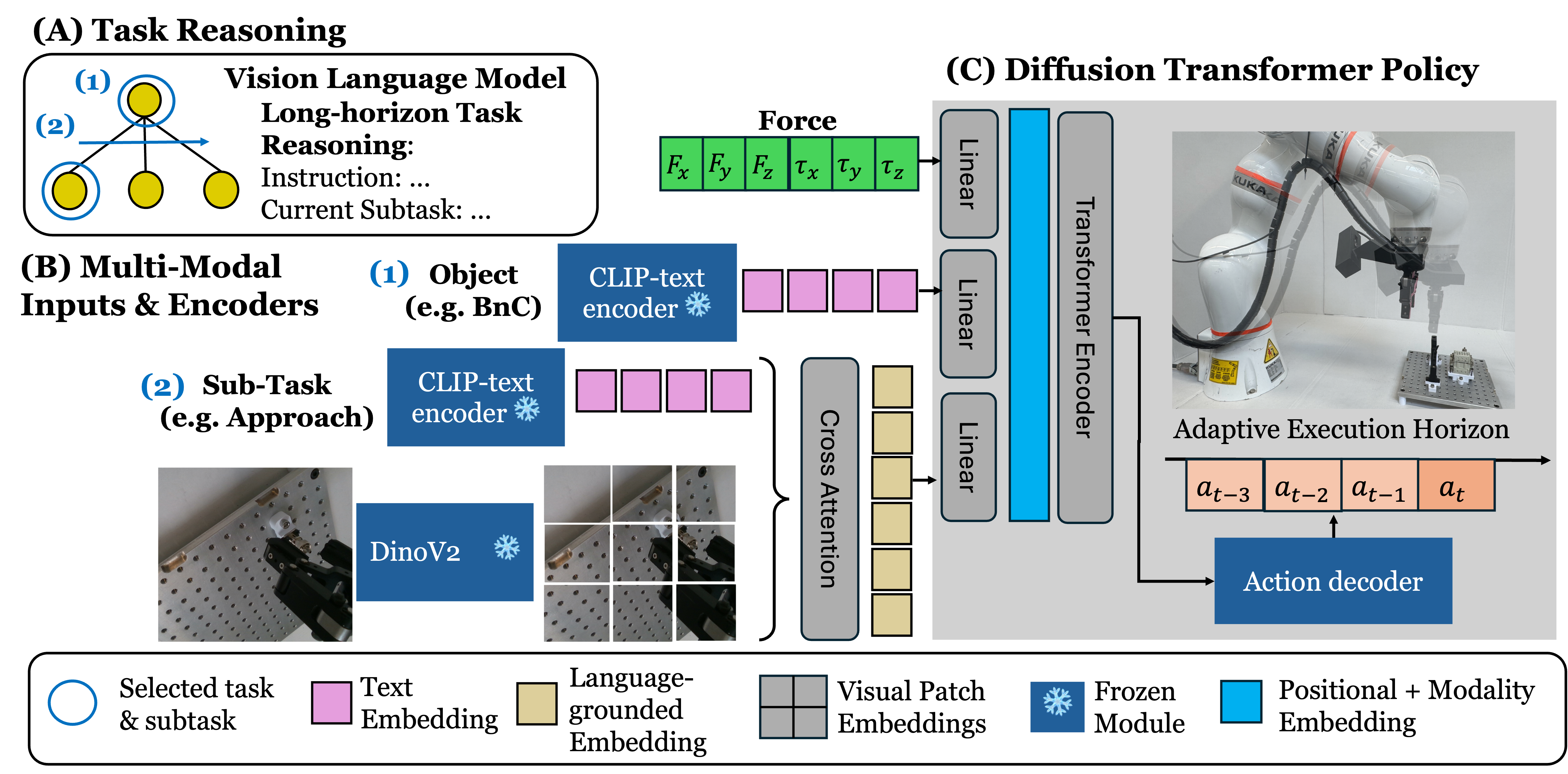}
    \caption{The VLM performs task selection from the human language input. The selected task description is encoded using a CLIP text encoder. Wrist-camera observations are processed by a DINOv2~\cite{oquab2023dinov2}, and the resulting image patches and language features undergo cross-attention to produce a task-relevant feature map. The DiT architecture then receives tokens corresponding to the attended visual features, subtask embeddings, and force inputs, each projected through a linear layer, enabling DiT to attend to multiple modalities during action prediction.}
    \label{fig:system_architecture}

\vspace{-1.5em}

\end{figure*}
\section{Related Work}
\label{Related Work}

\noindent \textbf{Language-Guided Imitation Learning}:
Behavior cloning provides an effective framework for learning manipulation skills from demonstrations \cite{Mandlekar2020LearningTG, florence2019self}, while diffusion-based policies have recently emerged as powerful generative models for complex robotic action distributions \cite{sohl2015deep, ho2020denoising, chi2023diffusionpolicy, chi2024diffusionpolicy, kang2025robotic}. In parallel, vision-language-action (VLA) models leverage pre-trained vision-language models (VLMs) to incorporate semantic task information into robotic policies, supporting object grounding and end-to-end action prediction \cite{Beyer2024PaliGemmaAV, kim2025openvla}. Representative models such as OpenVLA \cite{kim2025openvla} and $\pi_{0.5}$ \cite{Intelligence202505AV} demonstrate the effectiveness of large-scale vision-language representations for robotic manipulation.

\noindent \textbf{Hierarchical Learning and Planning for Robotic Manipulation}: Recent robotic manipulation systems have adopted language-model and VLM reasoning for hierarchical task planning and decision making \cite{ahn2022can,Zhao2025CoTVLAVC,huang2025thinkact,lin2025onetwovlaunifiedvisionlanguageactionmodel}. For example, CoT-VLA \cite{Zhao2025CoTVLAVC} generates intermediate subgoal images to guide manipulation, while ThinkAct \cite{huang2025thinkact} and OneTwoVLA \cite{lin2025onetwovlaunifiedvisionlanguageactionmodel} integrate reasoning and acting within a unified architecture. Hierarchical imitation learning and task segmentation have also been explored for long-horizon robotic manipulation \cite{ahn2022can,pmlr-v267-shi25d}. However, these methods are primarily evaluated on household manipulation tasks, with limited exploration of contact-rich assembly or disassembly. Furthermore, keypoint-based grounding methods \cite{fang2025kalm} are less applicable to industrial disassembly, where components are often small, visually similar, and out-of-distribution for general-purpose foundation models. Relatively few works study semantic task selection and language-guided object grounding for robotic disassembly without explicit object annotations.

\noindent \textbf{Counterfactual Learning for Robotic Manipulation}:
Counterfactual learning has been explored for causal representation learning, robust prediction, and generative modeling \cite{johansson2016counterfactual,kaushik2020counterfactual,sauer2021counterfactual}. In robot learning, MoCoDA \cite{pitis2022mocoda} and RoCoDA \cite{ameperosa2025rocoda} use counterfactual data augmentation to improve policy generalization, while Wang et al.~\cite{wang2024grounding} generate counterfactual failures to ground language plans in demonstrations. Relatedly, Ma et al.~\cite{ma2025contrastive} employ contrastive learning to improve language-conditioned representations for multi-task manipulation. However, these approaches do not address task-conditioned target disambiguation among multiple visually similar components in contact-rich manipulation. In contrast, our work incorporates counterfactual supervision into the diffusion objective to explicitly strengthen task--object grounding, without generating additional counterfactual trajectories.

\section{Problem Formulation}
\label{section: problem_formulation}

\noindent
Let us consider a scenario with $N$ disassembly tasks present in a single scene, where each parent task $\mathit{T}_p$ must be executed in order and is composed of a sequence of subtasks, $T_{s_i}$, each requiring a specific skill $\mathit{s}$. We define the parent task as a sequence of subtasks
$\mathit{T}_p = \{\mathit{T}_{s_i}\}_{i=1}^{n}$, where each subtask $\mathit{T}_{s_i}$ is associated with a required skill $\mathit{s}_i$. A robust policy must therefore identify and select the appropriate skill $\mathit{s}_i$ in order to successfully complete the overall $\mathit{T}_p$.

We aim to learn a policy $\pi$ that produces actions $\bm{a}_t \sim \pi(\bm{a}_t \mid \bm{O}_t)$ conditioned on multi-modal observations $\bm{O}_t = \{\bm{u}_t, \bm{I}_t, \bm{F}_t, \bm{l}_t\}$, where $\bm{u}_t$ denotes robot proprioception, $\bm{I}_t$ denotes visual input,
$\bm{F}_t$ represents force measurements, and $\bm{l}_t \in \mathcal{L}$ is a language descriptor specifying the task object $\mathit{o}$ and subtask skill $\mathit{s}$. The language input serves as a semantic conditioning variable that informs the policy about the current task. The policy should produce $\bm{a}_t$ consistent with the task specification $\bm{l}_t$, even under varying assembly configurations. We propose our language-grounded visual learning framework in Section~\ref{section: disambiguating}, language-conditioned policy learning in Section~\ref{section: language_policy}, and the automatic skill-selection framework in Section~\ref{section: online_skill}. The overall architecture is shown in Figure~\ref{fig:system_architecture}.

\noindent \textbf{VLM-Assisted Demonstration Trajectory Labeling:}
In prior work \cite{kangtaskcontext2025}, subtask labels for human demonstration data were generated using a heuristic approach based on robot end-effector position and gripper width. While this is effective in some cases, it does not generalize well when visual cues are essential for subtask segmentation. In such scenarios, manual labeling—where humans visually inspect and annotate the dataset—remains the most reliable method but is highly labor-intensive and costly.
To address this challenge, we use a vision-language-enhanced labeling framework that leverages VLMs as automated annotators. Specifically, we employ GPT-5 as the primary model for offline labeling.

\noindent \textbf{Diffusion Model Preliminaries:} Diffusion models are probabilistic generative frameworks that approximate the data distribution $p(x_0)$ through a gradual noising process over latent variables $\mathbf{x}_1, \mathbf{x}_2, \dots, \mathbf{x}_T$, 
which are progressively noisier versions of the clean data $\mathbf{x}_0$. Detailed discussions of the diffusion models and policy formulation can be found in \cite{kang2025robotic, kangtaskcontext2025} and \cite{chi2023diffusionpolicy, ho2020denoising}.

\begin{figure}[!t]
    \centering
    \includegraphics[width=\linewidth]{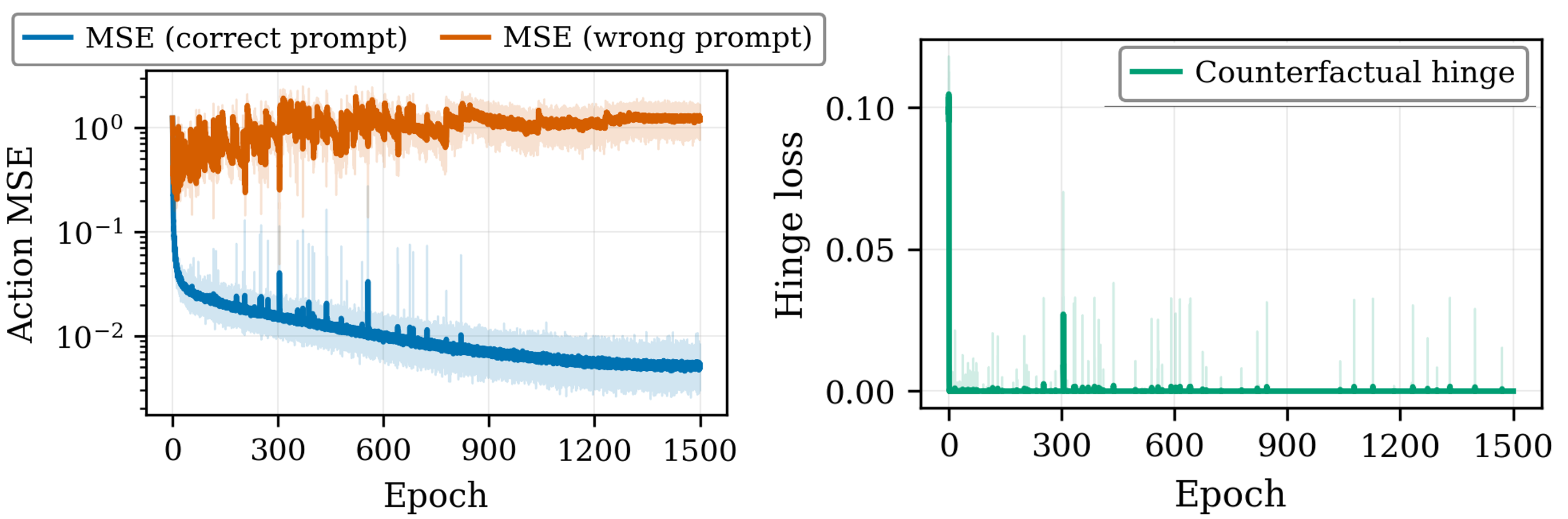}

    \caption{\textbf{CF loss behavior during training.}
The MSE for correct and mismatched action--language pairs diverges over training, while the CF hinge loss converges toward zero.}
    \label{fig:loss_plot}
\end{figure}

\section{Method}
\label{section: method}

\subsection{Disambiguating Task Target in Multi-task Scene with Target-object Grounding}
\label{section: disambiguating} 
\noindent Training a single policy across multiple tasks and targets can cause gradient interference, leading to incorrect object selection and mode averaging. Explicit object detection can alleviate this ambiguity but requires costly annotations and is challenging for small, diverse disassembly components. We therefore investigate whether language can ground the target object without object-level annotations.

Cross-attention has been widely adopted in vision-language models to align visual and language representations for downstream tasks \cite{kim2025openvla,alayrac2022flamingo, karamcheti2024prismatic}. While these representations generalize well to common objects, industrial disassembly involves small, visually similar components that are largely absent from large-scale vision-language training data. We therefore employ a cross-attention module to ground domain-specific task language in spatial DINOv2 patches, enabling the policy to associate specialized manipulation skills with the intended target object.

Additionally, we incorporate counterfactual supervision directly into diffusion-based policy learning. Specifically, we use counterfactual language labels to discourage actions conditioned on incorrect task-object associations while reinforcing actions corresponding to the intended target. At each denoising step, the policy is conditioned on an incorrect language command for the same action chunk, and its prediction is penalized unless it is worse than the prediction under the correct command. As shown in Figure~\ref{fig:loss_plot}, this objective causes the incorrect-command prediction error to diverge from the correct-command prediction error over training.

\begin{figure}[!t]
    \centering
    \includegraphics[width=\linewidth]{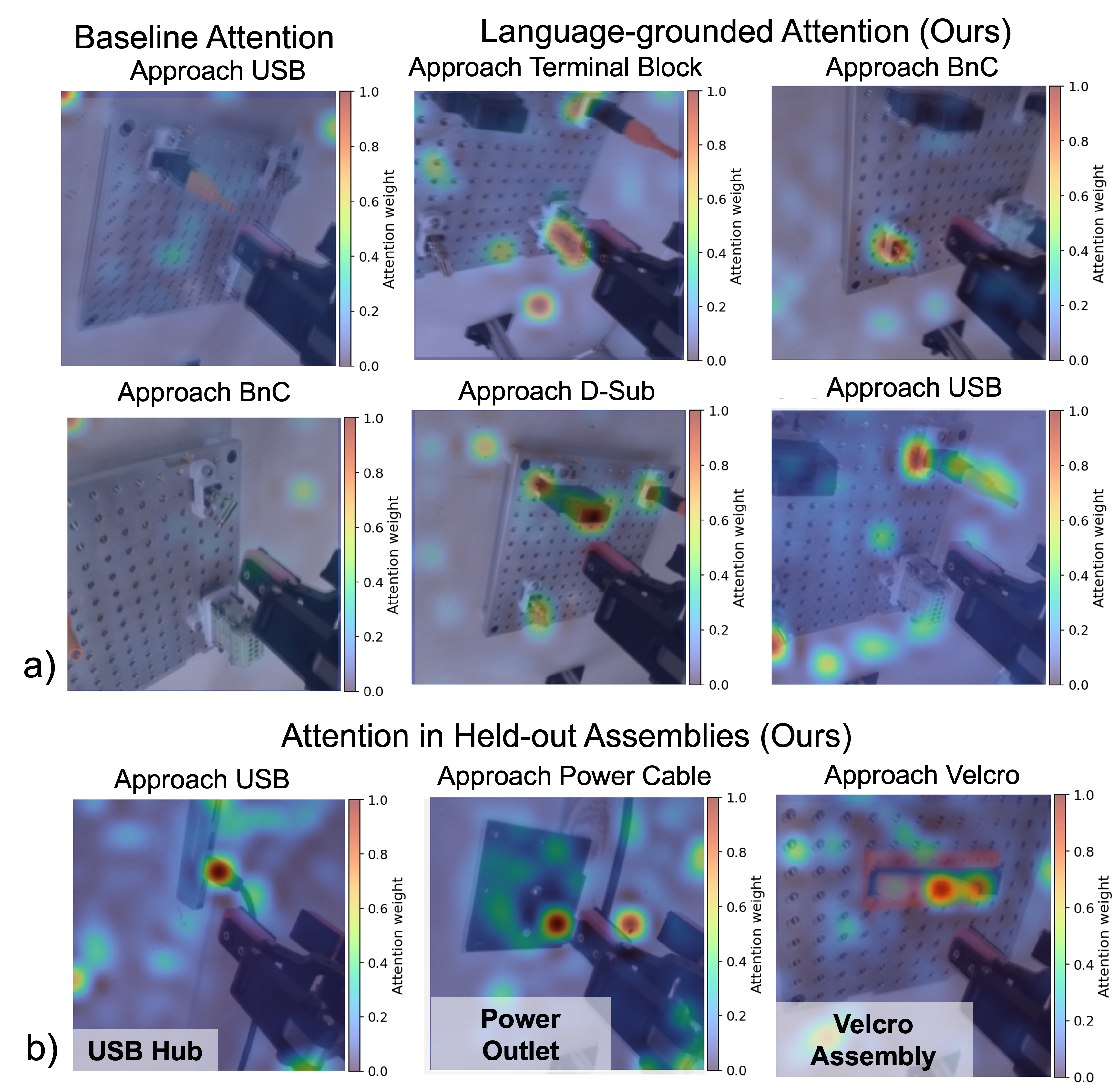}

\caption{(a) Attention maps during the approach phase under different language commands. Without the proposed grounding mechanism, DiT attends to seemingly unrelated regions. (b) Attention map for held-out assemblies.}
    \label{fig:atten_vis}
\end{figure}

\noindent \textbf{Standard MSE.} Let $\bm{c}^{+} = (\bm{u},\bm{F}_t,\bm{T}_s,\bm{I}_t,\bm{T}_o)$   \footnote{The parent task $T_p$ identifies the target object; we write it as $T_o$ in this section to emphasize the object stream that the counterfactual swaps.} denote the
correctly conditioned context. The noise prediction loss is
\begin{equation}
\mathcal{L}_{\text{mse}}(\bm{\theta};\bm{c}^{+})
=
\mathbb{E}_{k,\bm{\epsilon}}\!\left[
\bigl\|\bm{\epsilon}_{\bm{\theta}}(\bm{a}^{(k)},k,\bm{c}^{+}) - \bm{\epsilon}\bigr\|_2^2
\right]
\end{equation}

where $\bm{a}^{(k)}$ denotes the action at diffusion step $k$, $\bm{\epsilon}_{\bm{\theta}}$ is the predicted noise, $k$ is the diffusion timestep, and $\bm{\epsilon}$ is the ground-truth noise.
\noindent \textbf{Counterfactual hinge.}
Let the counterfactual context be
$\bm{c}^{-}_{o'} = (\bm{u},\bm{F}_t,\bm{T}_s,\bm{I}_t,\bm{T}_{o'})$, obtained by
swapping \emph{only} the object stream to a different class
$o'\!\neq\!o$ while leaving the subtask, image, force and proprioception
streams intact. The hinge term against $o'$ is
\begin{equation}
h_{o'}(\bm{\theta})
=
\bigl[\,\mathcal{L}_{\text{mse}}(\bm{\theta};\bm{c}^{+})
- \mathcal{L}_{\text{mse}}(\bm{\theta};\bm{c}^{-}_{o'}) + \delta\,\bigr]_{+}
\label{eq:hinge}
\end{equation}
with $[x]_{+}\!=\!\max(0,x)$ and margin $\delta{>}0$ ($\delta{=}0.1$).
We average over all $N_o{-}1$ wrong objects,
\begin{equation}
\mathcal{L}_{\text{cf}}(\bm{\theta})
\;=\;
\frac{1}{N_o-1}\sum_{o'\neq o} h_{o'}(\bm{\theta})
\label{eq:cf}
\end{equation}
The CF term is applied only on samples whose subtask $s$ lies in the
configured subset $\mathcal{S}_{\text{cf}}\!\subseteq\!\mathcal{S}$
(default $\mathcal{S}_{\text{cf}}{=}\{\text{Approach}\}$): later subtasks
are physically committed to the target and do not exhibit the
language-ignoring failure mode. Based on this, we make two remarks:
\textbf{R1. Hinge collapse certifies language use.} For an
input-ignoring policy $\bm{\epsilon}_{\bm{\theta}}(\cdot,\cdot,\bm{c}^{+})
\!\equiv\!\bm{\epsilon}_{\bm{\theta}}(\cdot,\cdot,\bm{c}^{-}_{o'})$,
the two MSE values in \eqref{eq:hinge} are equal, hence
$h_{o'}\!=\!\delta$ for every $o'$ and
$\mathcal{L}_{\text{cf}}\!=\!\delta$. The hinge can drop below $\delta$
\emph{only} if changing $\bm{T}_{o'}$ measurably degrades the prediction;
$\mathcal{L}_{\text{cf}}{\to}0$ therefore certifies that the object
stream actively steers $\bm{\epsilon}_{\bm{\theta}}$ along the dataset
trajectories. \textbf{R2. Margin breaks symmetry.} Without $\delta$, the trivial
solution $\bm{\epsilon}_{\bm{\theta}}\!=\!0$ in both branches satisfies
$h_{o'}\!=\!0$ trivially. The margin forces a strict gap
$\mathcal{L}_{\text{mse}}(\bm{c}^{-}_{o'}) \geq
\mathcal{L}_{\text{mse}}(\bm{c}^{+}) + \delta$, which is achievable
only when the language is informative. This framework enables the policy to attend to the appropriate target object, shown in Figure~\ref{fig:atten_vis}.
\vspace{-0.5em}

\subsection{Task-Context-Conditioned Diffusion Transformer Action Head}

\label{section: language_policy} 
\noindent We use a Diffusion Transformer (DiT) for action decoding with causal masking over both the noisy action sequence and encoder--decoder attention. Thus, each action prediction at horizon step $t$ attends only to action steps $\leq t$ and non-future state tokens, similar to \cite{chi2023diffusionpolicy}.

To address task ambiguity and the temporal freezing effect, we build upon the language-based task conditioning framework of~\cite{kangtaskcontext2025}. Unlike~\cite{kangtaskcontext2025}, which concatenates overall and subtask labels with visual observations, we represent the parent task $\mathit{T_p}$ separately and provide the subtask $\mathit{T_s}$ as a dedicated token to the diffusion transformer (Section~\ref{section: disambiguating}). The overall architecture of the proposed framework is shown in Figure~\ref{fig:system_architecture}.

\begin{figure}[tp]
    \centering
    \smallskip
    \includegraphics[width=\linewidth]{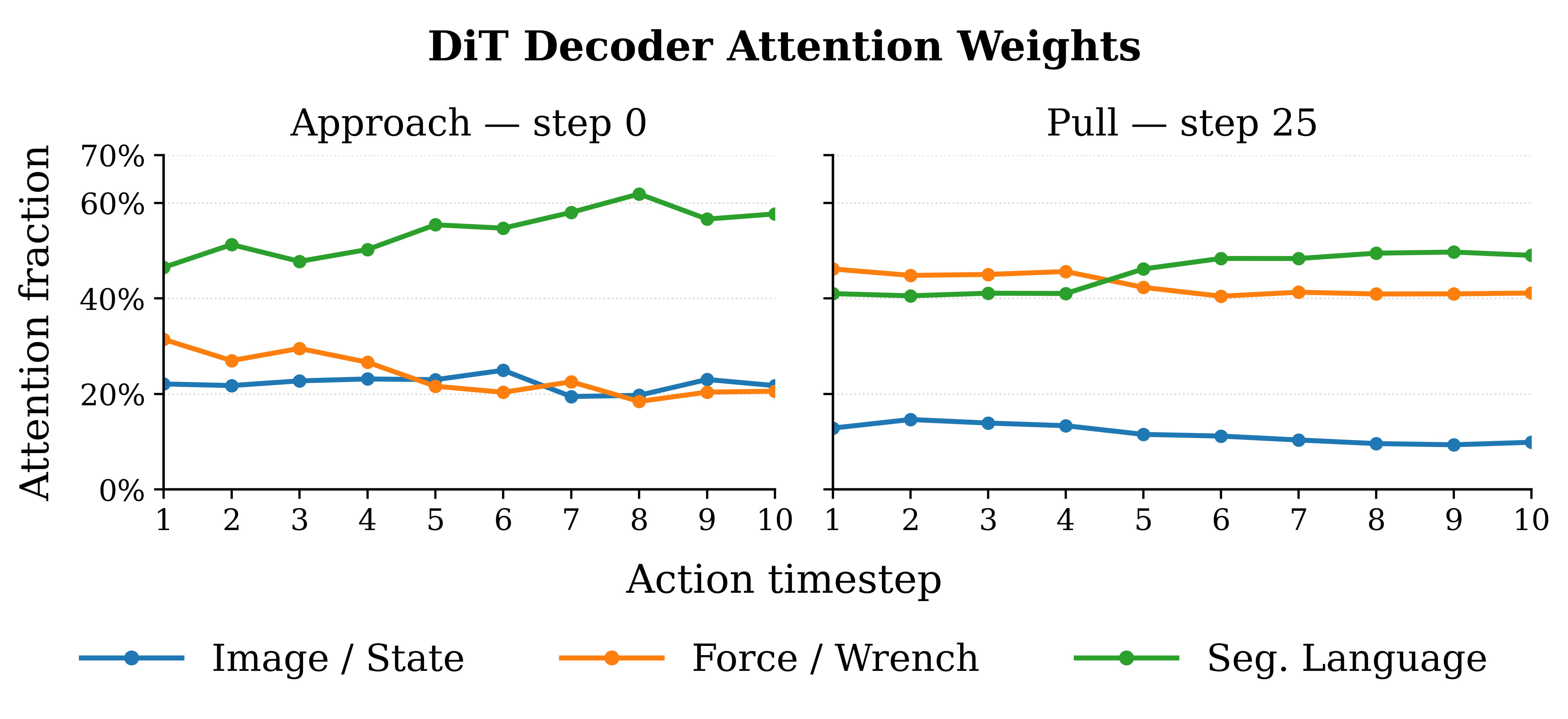}
    \vspace{-2em}
    \caption{Attention weights computed for different modalities during action prediction in the DiT policy. Compared to the approach phase, contact-rich tasks such as pulling exhibit higher attention weights toward force inputs.}
    \label{fig:atten_plot}

\end{figure}

In~\cite{kangtaskcontext2025}, the relative importance of vision and force modalities is manually specified for each subtask. In contrast, our architecture provides subtask and force representations as separate inputs to the action decoder, allowing cross-attention to learn their relative importance. Figure~\ref{fig:atten_plot} illustrates the resulting attention over different modalities during action generation.

\begin{figure}[!t]
    \vspace{2mm}
    \centering
    \includegraphics[width=0.8\linewidth]{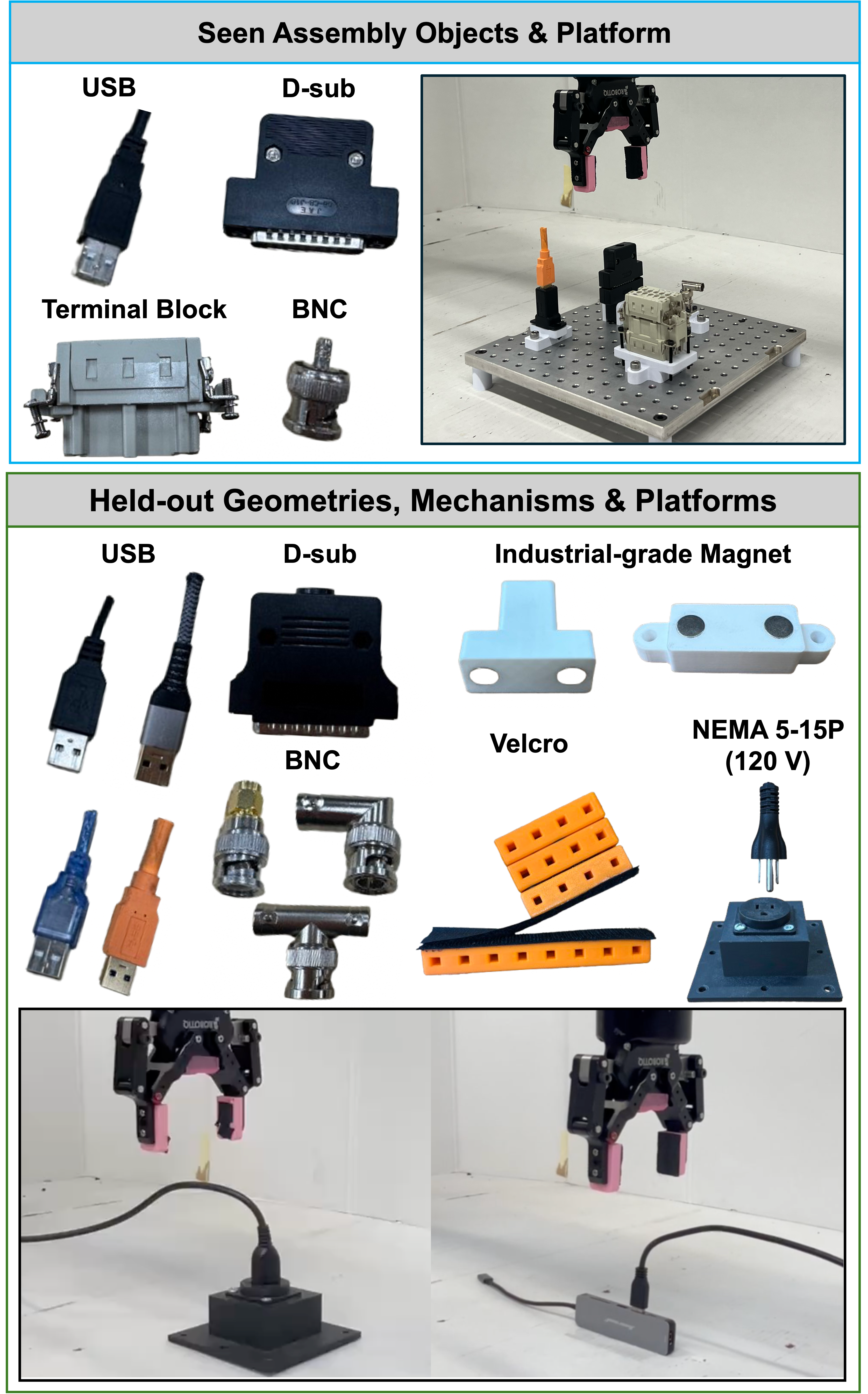}

    \caption{Diverse types of connectors and assembly configurations used to test generalizability of our approach.}
    \label{fig:parts}
\end{figure}

\subsection{Online Task Selection for Long-Horizon Execution}
\label{section: online_skill} 

\noindent Online subtask selection is required to autonomously switch task context during execution. The high-level planner is queried asynchronously from the low-level control policy at the highest rate permitted by its inference latency, while the low-level policy continues to operate at its fixed control frequency. Consequently, planning latency is amortized over multiple low-level policy executions and does not directly determine the policy control frequency. In this work, we propose a VLM-based task selector that enables semantic reasoning for subtask selection and generalization to previously unseen components while reusing the existing skill library. 

\noindent \textbf{VLM for Subtask Selection:} We leverage the semantic reasoning capabilities of VLMs for subtask selection and high-level planning. In our end-to-end experiments (Section~\ref{section:end_to_end_success}), we evaluate both a cloud-based GPT-5 planner and a locally deployed Qwen3-VL-8B planner to explore different deployment options. 
For both subtask labeling and online task selection, we find that augmenting the visual prompt with structured robot-state information substantially improves subtask prediction accuracy while naturally enabling failure recovery during online deployment. In addition to visual observations, the VLM receives force/torque measurements, gripper width, end-effector pose, and incremental end-effector motion encoded as natural language. 
\vspace{-0.5em}
\section{Experimental Setup}
\label{section: training} 
\subsection{Testbed}
\noindent
We evaluate our method on a KUKA LBR IIWA14 robotic arm.
Inspired by the NIST Task Board \#1 \cite{kimble2020benchmarking}, we design a custom testbed for data collection.
We conduct experiments under a variety of connection configurations and spatial arrangements, including scenarios in which two identical connectors coexist on the same board. 
The evaluation objects and assembly configurations are shown in Figure \ref{fig:parts}.

\subsection{Data Collection and Training}
\begin{figure}[tp]
    \centering
    \smallskip
    \includegraphics[width=0.9\linewidth]{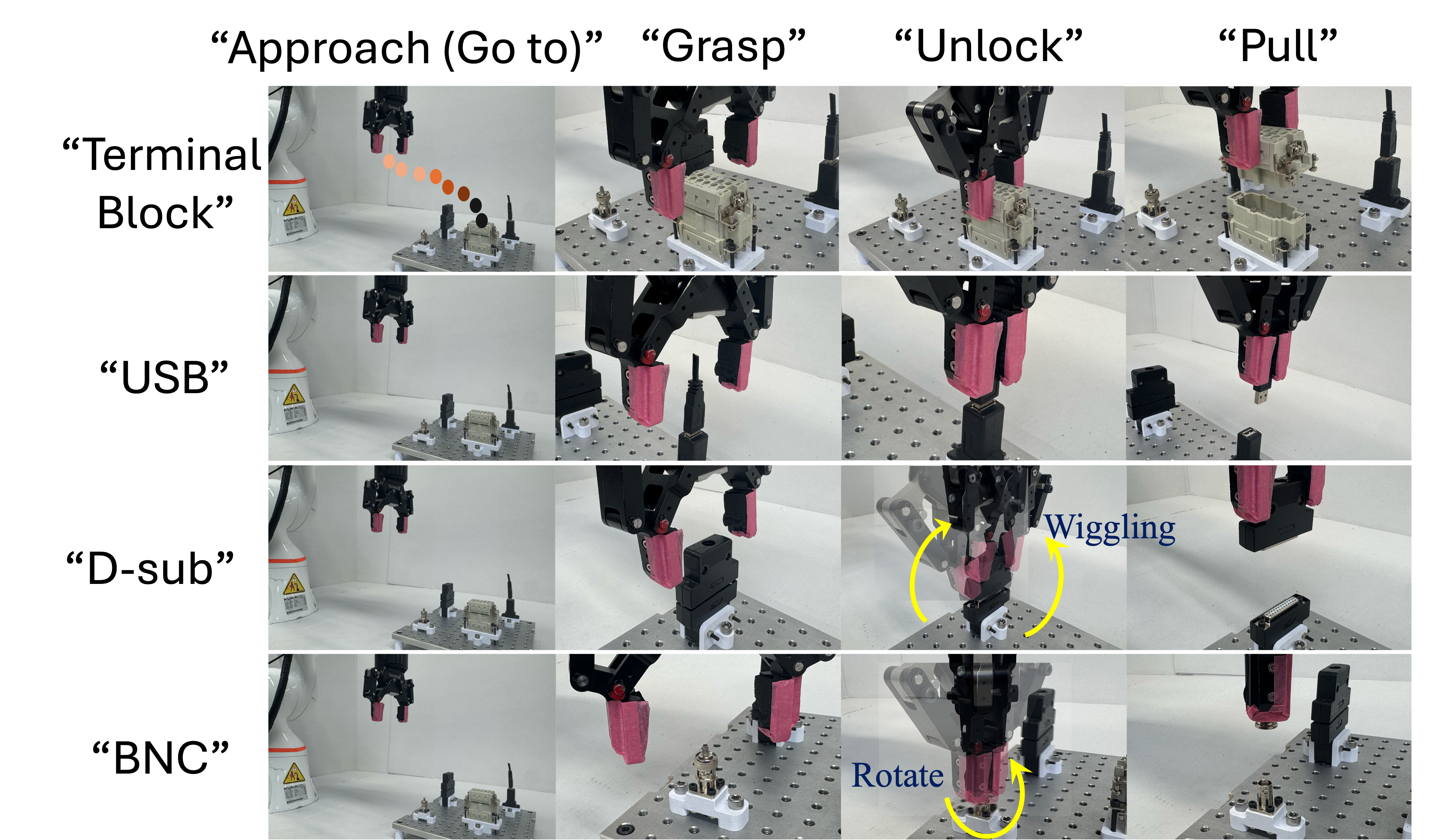}
    \caption{\textbf{Task Description in Text}: We illustrate a sequence of parent tasks and subtasks present in our scenario. A single policy executes different unlocking skills tailored to each mechanism.}
    \label{fig:trajectory_command}
\end{figure}

\noindent
We collect demonstrations via kinesthetic teaching following \cite{kang2025robotic, kangtaskcontext2025}. During data collection, we vary the connector’s position and surrounding configurations to promote generalization across diverse assembly layouts and connector placements. We use only the objects and assembly configurations marked as seen in Figure~\ref{fig:parts} for training, while the remaining objects are reserved for held-out evaluation. After collecting the trajectory data, we label each segment with corresponding language instructions following the hierarchical structure of parent tasks and subtasks shown in Figure \ref{fig:trajectory_command}.
Our training dataset contains 102 USB, 118 D-sub, 115 terminal-block, and 121 BNC demonstrations. Each demonstration consists of a complete trajectory, from the initial approach to connector extraction. We train our policy end-to-end using a CLIP text encoder and a single vision encoder (DINOv2) \cite{oquab2023dinov2} with frozen weights, as shown in Figure~\ref{fig:system_architecture}.
Training is conducted with a batch size of 128, learning rate of $1 \times 10^{-4}$, and weight decay of $1 \times 10^{-4}$. We apply random cropping and color jittering for data augmentation and train all models on a single NVIDIA RTX 4080 GPU for approximately 22 hours.

\section{Results}
\label{section:results} 
\noindent In the results section, we aim to answer four key questions regarding the effectiveness of the proposed framework. 
\textbf{(Q1)} Which VLM provides the most reliable performance for both offline subtask labeling and online task selection?
\textbf{(Q2)} How do different task-selection mechanisms, including heuristic-based, CNN-based, and VLM-based planners, compare in terms of end-to-end task success, robustness to held-out assembly configurations, and mitigation of the temporal freezing effect?
\textbf{(Q3)} Can the proposed method consistently identify the target object specified by the language command without requiring explicit object annotations?
\textbf{(Q4)} Can the proposed framework generalize to held-out connector mechanisms and held-out assembly configurations?

\begin{figure}[!t]
    \centering
    \includegraphics[width=0.9\linewidth]{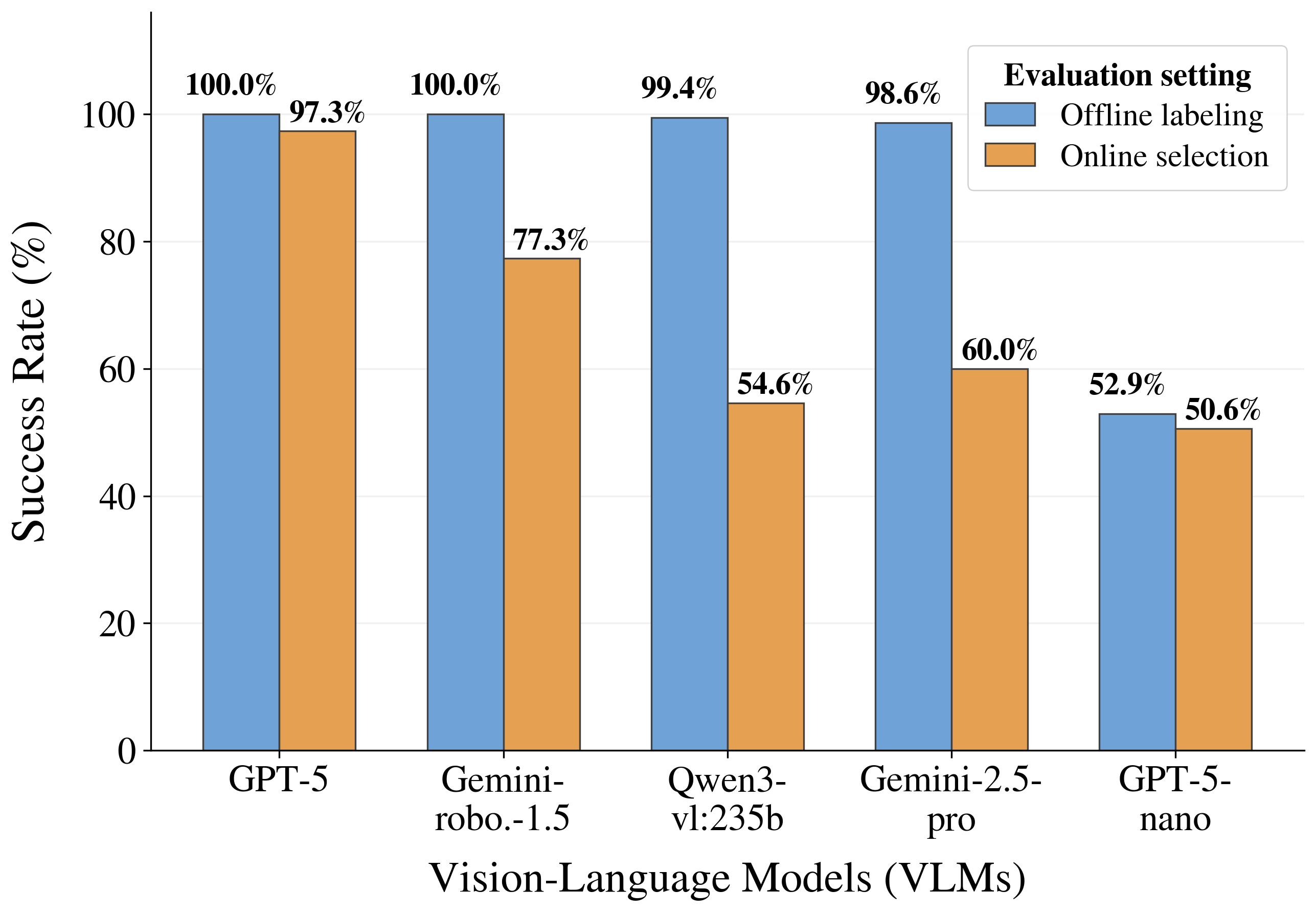}
    \caption{Comparison of VLM performance across two task-selection settings. Blue bars show offline trajectory-labeling accuracy, while orange bars show online next-subtask selection accuracy during robot execution.}
    \label{fig:VLM_eval}
\end{figure}

\subsection{Evaluation of Vision--Language Models for Task Selection}
\label{section:vlm_eval}

\noindent We evaluate five VLMs for two task-selection capabilities: (i) offline labeling of human demonstration trajectories and (ii) online proposal of the next subtask during robot execution. For offline labeling, we quantitatively evaluate accuracy by comparing VLM predictions against human-annotated ground-truth labels. For online evaluation, we assess whether the proposed subtask is consistent with the current robot state and task progression. Results for both offline labeling and online subtask selection are shown in Figure~\ref{fig:VLM_eval}.

For trajectory labeling, we evaluate 10 trajectories across four connector types, yielding 694 labeled samples.  Larger models generally achieve higher labeling accuracy, although at the cost of increased inference latency. Since offline labeling does not require real-time execution, accuracy is more important than latency for this component. We therefore use GPT-5 for offline labeling due to its superior accuracy, while noting that Gemini Robotics-ER 1.5 and Gemini-2.5-Pro achieve competitive performance with lower latency. 

We further evaluate each VLM's ability to propose the next subtask during online execution. This setting is more challenging than offline labeling because the model must infer the next action from the current scene and robot state. We evaluate 25 trajectories, producing 150 proposal queries. As shown in Figure~\ref{fig:VLM_eval}, GPT-5 achieves the highest online proposal accuracy (97.3\%). The Gemini models commonly fail by predicting ``done'' too early or too late, while Qwen3-VL-235B (Remote) struggles to transition to the pulling phase. GPT-5-nano often remains in the ``approach'' phase and fails to advance to the next subtask.

\subsection{End-to-End Task Success Rate Comparison}
\label{section:end_to_end_success}
\noindent To assess the contribution of each component in our framework, we benchmark against the following policy variants:
\begin{enumerate}
    \item \textbf{DP-S (Single-Task Diffusion Policy~\cite{chi2023diffusionpolicy}):}
    A diffusion policy trained independently for each task, without shared representations or multi-task conditioning.

    \item \textbf{DP-M-C-CNN (Multi-Task CNN-Based Diffusion Policy~\cite{kangtaskcontext2025}):}
    A task-context-aware diffusion policy from prior work that uses a CNN-based backbone and conditions the policy on a flattened language-command embedding.

    \item \textbf{DP-M-T-CF-RB (Multi-Task Transformer with Rule-Based Selection):}
    A transformer-based multi-task diffusion policy trained with CF loss for improved language-to-object grounding with subtask detection using hand-crafted rules based on robot state and force signals.

    \item \textbf{DP-M-T-CF-CNN:}
    A transformer-based multi-task policy with lightweight CNN-based subtask selection.
    
    \item \textbf{DP-M-T-CF-VLM (Ours, Multi-Task Transformer with CF Loss and VLM-Based Selection):}
    Transformer-based multi-task diffusion policy with counterfactual (CF) loss, cross-attention, and VLM-based online subtask selection. Two variants are evaluated: \textbf{V1}, which uses a locally deployed Qwen3-VL-8B planner, and \textbf{V2}, which uses a cloud-hosted GPT-5 planner.

\end{enumerate}

Each experiment is repeated 20 times per connector type. We define success as the robot completing the full disassembly by removing the connector within a limit of 50 action steps. For multi-task policies, we additionally require correct selection of the parent task $T_p$ based on the provided language instruction; initiating an incorrect task is counted as a failure.

During the experiments, we evaluate generalization across spatial arrangements, multiple identical connectors not observed during training, and variations in connector geometry, as shown in Figure~\ref{fig:parts}. Additional evaluations on held-out connector mechanisms and assembly configurations are conducted only for CNN and VLM-V2 planners, as the remaining methods did not generalize to these larger distribution shifts in preliminary trials. These results are presented in Section~\ref{section:cnn_vs_vlm}.

The baselines DP-S and DP-M-C-CNN compare our framework against existing methods. The remaining three methods evaluate different subtask planning strategies while sharing the same diffusion policy. The rule-based planner determines the next subtask using thresholds on gripper width and end-effector height. Results are summarized in Table~\ref{tab:connector_disassembly}. We see that compared to existing methods, our method performs substantially better due to better multi-modal weighting (Figure~\ref{fig:atten_plot}) during inference and language-to-object grounding that enables robust target object selection. Particularly, DP-M-C-CNN's low success rate stems from its flattened language embedding, which provides no spatial correspondence to the visual scene and thus fails to ground the command to the correct target when similar connectors coexist. Note that the framework in \cite{kangtaskcontext2025} was designed for a fixed assembly board, so these results reflect its behavior under the more diverse configurations considered here, although it was trained on the same demonstration data as the proposed method.

The V1 and V2 benchmarks compare locally deployed and cloud-based VLM planners.  We use Qwen3-VL-8B as an open-source alternative deployable on local hardware, which achieves 53.3\% online subtask prediction accuracy on the same query set used in Section~\ref{section:vlm_eval}. The GPT-5 planner achieves higher end-to-end task success due to its more reliable semantic task predictions. Under our experimental setup, GPT-5 produces one task prediction every $2.9  \pm 1.1$ policy action steps and Qwen3-VL-8B one every $3.7 \pm 0.6$ steps, measured over $250$ VLM calls. We report planner latency in units of the low-level control loop, since absolute response times for cloud-hosted models conflate network transit with inference cost. Notably, the cloud-hosted GPT-5 planner sustains a higher update rate than Qwen3-VL-8B running locally on a single RTX 3080 (used for inference), indicating that on-premises deployment of an 8B model on consumer hardware does not necessarily reduce planner latency. Additionally, it is worth noting that incorrect subtask predictions do not necessarily cause immediate task failure because the diffusion policy additionally conditions on visual observations and robot state (Figure~\ref{fig:atten_plot}). Nevertheless, more accurate task predictions provide more consistent task context, reducing policy drift during long-horizon execution.

\noindent \textbf{Common failure modes highlight the importance of reliable task selection.} Misalignment and incorrect target selection are the primary causes of failure. Among the evaluated planners, the heuristic rule-based method is particularly susceptible to diverse trajectories, often triggering subtasks prematurely or remaining stuck on the current subtask.

\subsection{CNN vs. VLM Planner Study}
\label{section:cnn_vs_vlm}
To examine when a lightweight learned task classifier is sufficient and when broader semantic reasoning is beneficial, we compare the proposed VLM-based planner with a CNN-based alternative. The CNN predicts the next subtask from short sequences of multimodal observations, including RGB images, force/torque measurements, and robot state, using a fine-tuned ResNet34 followed by a temporal Conv1D classifier. We evaluate both planners across in-distribution and held-out settings to characterize their trade-offs in prediction accuracy, inference latency, and generalization.
    
\noindent \textbf{Accuracy and Latency:} The lightweight CNN-based task selector provides substantially lower inference latency than the VLM-based selector while achieving 92\% subtask prediction accuracy for in-distribution objects, often exhibiting a one- or two-step delay in predicting task transitions, particularly for data-scarce subtasks such as ``Pull.'' Due to its lower inference latency, the CNN-based planner can be queried at the same frequency as the action policy. In contrast, the VLM-based planner is queried less frequently. In-distribution, the two planners perform comparably (0.93 vs.\ 0.94), indicating that the CNN is sufficient when object categories and task transitions are known. The planners diverge under distribution shift: on held-out connector mechanisms the VLM-based planner outperforms the CNN by 48 percentage points on average (Table~\ref{tab:cnn_vlm_ood}), reflecting its ability to reason semantically rather than classify among trained categories.

\noindent\textbf{Semantic Planning Beyond Learned Task Classification.}
While the CNN-based planner provides efficient subtask prediction for predefined tasks, it is inherently limited to object categories and task transitions observed during training. In contrast, the VLM reasons jointly over visual observations and structured robot-state information, enabling semantic planning beyond closed-world task classification. We demonstrate two representative capabilities.

\noindent \textit{Held-out connectors:} We further evaluate the VLM-V2 and CNN-based planners on held-out connector mechanisms, including an industrial-grade magnetic connector, Velcro, a NEMA 5-15P 120 V power connector, and a USB hub (Figure~\ref{fig:parts}), none of which were included during training but all of which can be manipulated using the existing skill library. The power connector and USB hub are particularly challenging because they are mounted on previously unseen assembly platforms. The results are shown in Table~\ref{tab:cnn_vlm_ood}. The observed failures mainly occurred during gripper alignment in the approach phase because of the highly out-of-distribution connector geometries rather than incorrect task selection. Although connectors requiring entirely new manipulation skills remain outside the scope of this work, these results demonstrate the VLM's ability to generalize task selection across diverse connector geometries, locking mechanisms, and visual appearances while reusing the existing skill library.

It is worth noting that generalization to held-out connector mechanisms and assembly configurations arises from the combination of robust visual representations, language-conditioned task grounding even on unseen connectors (Figure~\ref{fig:atten_vis}(b)), and VLM-provided task labels, which together encourage the policy to attend to task-relevant objects rather than memorizing object identities or scene configurations. Therefore, we find the CNN-based planner better suited to closed-world, latency-sensitive deployment settings with known objects and skills, while the VLM-based planner is preferable when semantic generalization and failure-aware reasoning are needed.

\noindent \textit{Failure recovery:} Our task decomposition and VLM-based reasoning naturally support failure recovery. When failures occur, the VLM reasons over visual observations and structured robot-state information to determine that task execution has deviated from the expected progression. Representative cues include a fully closed gripper indicating a failed grasp or an end-effector pose unexpectedly close to the table. The planner then invokes a recovery primitive that returns the manipulator to the previous pose before replanning. Across ten induced failure cases, the VLM consistently identified the failure condition and selected the recovery primitive. While the proposed task decomposition also enables predefined heuristic-based failure recovery for known failure conditions, such approaches rely on manually specified rules and do not naturally infer when execution should revert to a previous subtask. In contrast, the VLM performs semantic reasoning over both visual observations and robot-state information, allowing it to determine the appropriate recovery stage without explicitly engineered transition rules.

\vspace{-1.0em}

\subsection{Object Grounding Ablation}
\label{section:ablation_cf_ca}

\noindent We evaluate the importance of our language-to-object grounding mechanism by ablating the CF loss and cross-attention architecture. All variants use the same frozen DINOv2 and CLIP encoders, so differences isolate the grounding mechanism rather than the choice of pretrained representation. As shown in Table~\ref{tab:cf-ablation}, object-selection performance improves progressively with counterfactual supervision and cross-attention, indicating that the gains arise from the proposed grounding mechanism rather than the pretrained DINOv2 backbone alone. Without these components, the policy may collapse to trajectories between objects or execute motions that match the commanded skill while grounding to the wrong target.

\begin{table}[!ht]
    \vspace{2.5mm}
    \centering
    \footnotesize
    \renewcommand{\arraystretch}{1.1} 
    \resizebox{\linewidth}{!}{ 
    \begin{tabular}{@{}lccccc@{}}
        \toprule
        & D-Sub 
        & USB 
        & BNC 
        & Terminal 
        & Avg. \\ 
        \midrule
        DP-S \cite{chi2023diffusionpolicy} & 0.65 & 0.60 & 0.45 & 0.65 & 0.59 \\ 
        DP-M-C-CNN \cite{kangtaskcontext2025} & 0.10 & 0.20 & 0.00 & 0.45 & 0.19 \\ 
        DP-M-T-CF-RB  & 0.45 & 0.65 & 0.55 & 0.95 & 0.65 \\ 
        
        DP-M-T-CF-CNN & 0.95 & 1.00 & 0.75 & 1.00 & 0.93 \\ 
        
        DP-M-T-CF-VLM-V1 & 0.85 & 0.90 & 0.60 & 0.90 & 0.81 \\ 
                
        \textbf{DP-M-T-CF-VLM-V2} & \textbf{0.95} & \textbf{1.00} & \textbf{0.80} & \textbf{1.00} & \textbf{0.94} \\ 
        \bottomrule
    \end{tabular}
    }
    \caption{\textbf{Success Rate Comparison}: Each entry is the success rate over 20 trials per connector; our best-performing variant is shown in bold. Among the connectors, the BNC is the most challenging due to its small size and strong dependence on accurate subtask transitions.}
    \label{tab:connector_disassembly}
    \vspace{-0.5em}
\end{table}

\subsection{Number of Execution Prediction Steps Required: Regular Policy vs. Task-Context-Informed Policy}
\label{section:mode_switch_comparison}
\begin{figure}[!t]
    \centering
   \includegraphics[width=0.9\linewidth]{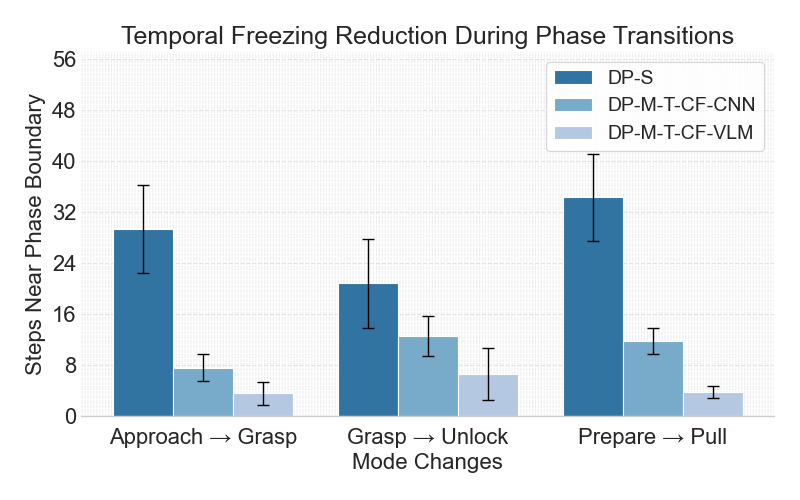}

    \caption{Action steps required for mode switching. Method names correspond to the benchmark variants defined in Section~\ref{section:end_to_end_success}.}
    \label{fig:mode_switch_comp}
\end{figure}

\noindent To evaluate whether our framework mitigates the temporal freezing bottleneck, we measure the number of action steps required for the policy to transition between tasks. We compare three methods: a baseline policy without subtask labels, a CNN-based classifier, and a VLM-based classifier. As shown in Figure~\ref{fig:mode_switch_comp}, the baseline policy tends to remain in idle states before committing to the next task, requiring the longest transition time and resulting in prolonged idle behavior. In comparison, both classifier-based methods reduce the number of action steps required for task switching. The CNN-based method requires slightly more action steps than the VLM-based method, while the VLM-based method produces the fewest delays in predicting the next subtask.

\begin{table}[t]
\centering
\footnotesize
\renewcommand{\arraystretch}{1.1}
\resizebox{\linewidth}{!}{%
\begin{tabular}{lcccc}
\toprule
Method & USB Hub & Power Connector & Magnetic & Velcro \\
\midrule
CNN    & 0.40 & 0.50 & 0.30 & 0.00 \\
\textbf{VLM-V2} & \textbf{0.90} & \textbf{0.90} & \textbf{0.70} & \textbf{0.60} \\
\bottomrule
\end{tabular}%
}
\caption{Task success rates on held-out connector mechanisms and configurations. Each experiment is repeated 10 times per connector.}
\label{tab:cnn_vlm_ood}
\end{table}

\begin{table}[t]
\vspace{2.5mm}

\centering
\renewcommand{\arraystretch}{1.1}
\begin{tabular}{lc}
\toprule
\textbf{Model} & \textbf{Successes / Trials} \\
\midrule
DiT (no Cross-Attn, no CF)   &           13/30\\
DiT + CF Loss           & 20/30 \\
DiT + CF Loss + Cross-Attn (Ours)       & 30/30 \\

\bottomrule
\end{tabular}
\caption{Correct object selection over 30 trials, given a language command from the user or the VLM.}
\label{tab:cf-ablation}

\end{table}

\subsection{Key Findings}

\noindent \textbf{Limitations of Behavior Cloning Methods.}
Conventional behavior cloning can overfit to specific assembly configurations and lacks explicit task structure, leading to poor generalization and temporal freezing in multi-task disassembly.

\noindent \textbf{VLM-Based Task Planning Improves Generalization.}
VLM-based planning improves task selection across held-out assemblies and connector mechanisms while supporting failure recovery through joint visual and robot-state reasoning.

\noindent \textbf{Task-context-aware grounding enables robust target selection.}
The combination of language-conditioned visual grounding and counterfactual supervision enables the policy to consistently identify task-relevant manipulation targets without requiring explicit object annotations.

\section{Conclusions}
\label{section:conclusions}
\noindent In this work, we present a framework for robotic disassembly that combines language-guided policy learning with autonomous hierarchical task selection. By combining language grounding for target-object disambiguation with VLM-based hierarchical task selection, the proposed framework improves the robustness of long-horizon robotic disassembly across diverse assembly configurations. Experimental results demonstrate that the combination of VLM-based semantic task planning and task-context-aware language grounding enables robust task selection, improves end-to-end task success, and generalizes across diverse assembly configurations and held-out connector mechanisms. However, the current approach relies on carefully designed prompts and close-range visual inputs, partly due to the limited visual reasoning capability of existing VLMs. Future work will investigate larger skill libraries and how policies can generalize to novel language commands for autonomous skill discovery.

\bibliographystyle{IEEEtran}
\bibliography{main}

\end{document}